\documentclass[10pt,twocolumn,letterpaper]{article}

\usepackage{cvpr}              

\usepackage{graphicx}
\usepackage{amsmath}
\usepackage{amssymb}
\usepackage{booktabs}
\usepackage{bigstrut}
\usepackage{multirow}
\usepackage{adjustbox}

\newlength\savewidth\newcommand\shline{\noalign{\global\savewidth\arrayrulewidth
  \global\arrayrulewidth 1pt}\hline\noalign{\global\arrayrulewidth\savewidth}}

\usepackage[pagebackref,breaklinks,colorlinks]{hyperref}

\providecommand{\ie}{\textit{i.e.}\@\xspace}

\usepackage[capitalize]{cleveref}
\crefname{section}{Sec.}{Secs.}
\Crefname{section}{Section}{Sections}
\Crefname{table}{Table}{Tables}
\crefname{table}{Tab.}{Tabs.}

\def\cvprPaperID{6984} 
\def\confName{CVPR}
\def\confYear{2023}

\begin{document}

\title{3D Point Cloud Pre-training with Knowledge Distillation from 2D Images}

\def\authorBlock{
    Yuan Yao$^1$~~~
    Yuanhan Zhang$^2$~~~
    Zhenfei Yin$^3$~~~
    Jiebo Luo$^1$~~~
    Wanli Ouyang$^3$~~~
    Xiaoshui Huang$^3$  \\
    $^1$~University of Rochester~~~~~$^2$~Nanyang Technological University\\$^3$~Shanghai AI Laboratory\\
}

\author{\authorBlock}


\maketitle

\begin{abstract}
   The recent success of pre-trained 2D vision models is mostly attributable to learning from large-scale datasets. However, compared with 2D image datasets, the current pre-training data of 3D point cloud is limited. To overcome this limitation, we propose a knowledge distillation method for 3D point cloud pre-trained models to acquire knowledge directly from the 2D representation learning model, particularly the image encoder of CLIP, through concept alignment. Specifically, we introduce a cross-attention mechanism to extract concept features from 3D point cloud and compare them with the semantic information from 2D images. In this scheme, the point cloud pre-trained models learn directly from rich information contained in 2D teacher models. Extensive experiments demonstrate that the proposed knowledge distillation scheme achieves higher accuracy than the state-of-the-art 3D pre-training methods for synthetic and real-world datasets on downstream tasks, including object classification, object detection, semantic segmentation, and part segmentation.
\end{abstract}

\section{Introduction}

In recent years, tremendous successes have been made in pre-training models \cite{radford2021learning,dosovitskiy2020image,liu2021swin,chowdhery2022palm,alayrac2022flamingo}. 
Transferring the rich knowledge of these models---as a starting point~\cite{zhou2021learning} or extracted feature~\cite{kolesnikov2020big}---can boost the performance of various downstream tasks in the field of computer vision~\cite{dosovitskiy2020image,liu2021swin} and natural language processing~\cite{chowdhery2022palm}. 

The fuel of large-scale pre-trained models relies on the availability of increasingly large and diverse datasets~\cite{deng2009imagenet,kuznetsova2020open}. However, compared with collecting large-scale 2D dataset---such as images~\cite{deng2009imagenet}, image-text pairs~\cite{radford2021learning}---collecting large-scale 3D datasets, such as point clouds~\cite{chang2015shapenet}, is non-trivial. The size of the modern 3D dataset is thus much smaller than 2D dataset---ShapeNet~\cite{chang2015shapenet}, which is adopted by the state-of-the-art~(SoTA) 3D pre-trained model, \ie Point-MAE~\cite{liu2022masked}, covers about 50k synthetic images. WIT~\cite{radford2021learning}, which the SoTA 2D pre-trained model adopts, \ie CLIP~\cite{radford2021learning}, covers about 400M image-text pairs. As a result, though CLIP improves the ``from-scratch'' model by 7.6 points on Caltech101~\cite{Caltech101,radford2021learning,zhai2019large},  Point-MAE~\cite{liu2022masked} only raise the mAP25 score by less than one point on the ScanNetV2~\cite{dai2017scannet}.
Witnessing the tremendous performance gain brought by large-scale 2D data, we seek to answer the following critical question - \textit{Can we elevate the 3D pre-trained models by unleashing the power of large-scale and easy-to-collect 2D data?}
%
%

\begin{figure}
    \centering
    \includegraphics[width=\linewidth]{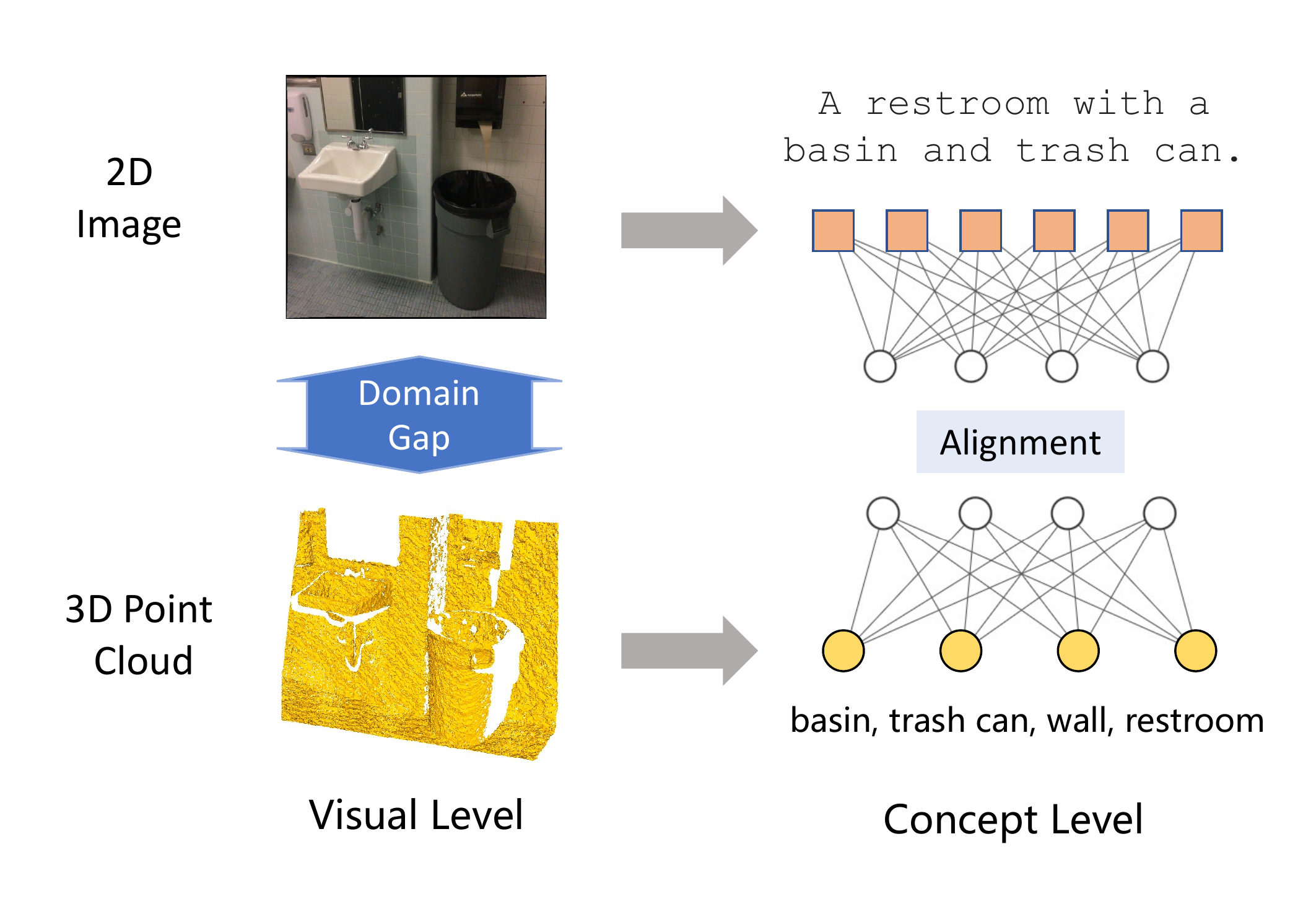}
    \caption{\textbf{Illustration of Concept Alignment.} Domain gap exists between 2D images and point clouds, which is an obstacle for 3D point cloud knowledge distillation from 2D images. However, 2D images and point cloud share semantic representations if they depict the same instance. Therefore, we propose to align the semantic prefix embeddings extracted from a 2D image with the concept tokens extracted from a point cloud.}
    \label{fig:illustration}
\end{figure}

In this work, we introduce a novel knowledge distillation framework to incorporate knowledge from the 2D domain into an effective 3D point cloud backbone transformer. More specifically, we observe that a large domain gap exists between 2D and 3D modalities.
To tackle this problem, we propose concept alignment across domains for knowledge distillation. 
Figure~\ref{fig:illustration} illustrates the key idea of concept alignment.
The intuition is that the semantic features extracted from 2D images are beneficial for the pre-training task of the point cloud domain, since 2D images and point cloud should share semantic representations if they depict the same instance, despite the difference in modality. 
We develop two independent branches for images and point clouds to extract the global semantic features from these two domains. For 2D semantic encoding, we introduce the Clip Caption model to gather semantic features~\cite{mokady2021clipcap}.
In the point cloud domain, inspired by the concept tokenizer framework~\cite{yang2022visual}, a cross-attention module is designed to extract tokenized concept features from the 3D point cloud.
The semantic features from the image domain and concept features from the point cloud domain are contrasted for distillation.

Experiments show that our approach is effective on downstream tasks, especially for object detection and semantic segmentation tasks on real-world datasets. For the ScanNetV2 dataset, we achieve 63.4 mAP25 in object detection and 46.4 mIOU in segmentation task, outperforming the state-of-the-art Point-MAE baseline by 1.7 points and 1.0 points respectively.

Our main contributions are as follows:
\begin{itemize}
\item We propose a knowledge distillation method for point cloud pre-training. The proposed concept extraction module and concept alignment mechanism distill knowledge based on semantic features, thus narrowing the domain gap between 2D and 3D domains.
\item Comprehensive experiments on both real-world and synthetic datasets demonstrate that knowledge distillation achieves better performance than the state-of-the-art 3D pre-training methods in various downstream tasks, such as object detection, semantic segmentation, object classification and part segmentation.
\item From the perspective of knowledge distillation, our work is the first to pre-train a point cloud backbone by directly learning from 2D representations. We demonstrate the possibility to address the domain gap between 2D and 3D, so that 3D pre-trained models can benefit from hyper-scale 2D datasets. We hope this work will advance the research of cross-domain learning between 2D and 3D domains.
\end{itemize}





\section{Related Work}
This section will review the related works from two aspects: 3D pre-training  and knowledge distillation.

\subsection{3D pre-training}
3D pre-training methods aim to learn a good initialization for the 3D backbone, which helps  3D representation learning. The good representation can then benefit 3D downstream tasks. The current existing 3D pre-training methods can be roughly divided into two categories: contrastive learning and reconstruction.

\textbf{Contrastive learning methods} pre-train the backbone by comparing the local or global structure information similarity from given point clouds. Typical examples are  Pointcontrast \cite{xie2020pointcontrast} and Crosspoint \cite{afham2022crosspoint}  methods. PointContrast compares the local point similarity to learn the  local point features extraction ability for the backbone.  Crosspoint compares the similarity between image global feature and point cloud global feature to pre-train the backbone. The pre-trained backbones can be utilized to solve 3D downstream tasks, such as point cloud detection, segmentation, recognition and registration.

\textbf{Reconstruction methods} encode the given point cloud into feature and decode a predicted point cloud from the feature. Next, the difference between the predicted and input point clouds are regarded as the loss to train the backbone \cite{wang2021unsupervised,zhang2021self,sun2022unsupervised}. Recently, the masking strategy has been adopted into the point cloud field and several masking based reconstruction methods are proposed to pre-train the backbone\cite{liu2022masked,yu2022point,pang2022masked}. They usually follow the pipeline of farthest point sampling (FPS) and multi-layer perception (MLP) to convert the given point cloud into a sequence of tokens. The masking strategy is then applied to these tokens and extracts deep features. Afterwards, the point cloud is reconstructed by using the deep features.

However, the above 3D pre-training methods require training on 3D datasets while  the current available data is limited. It is challenging to learn a pre-trained model of broad knowledge on a limited dataset. We propose a knowledge distillation method to inherit knowledge by leveraging 2D pre-trained models. Our method belongs to the category of 3D pre-training methods.

\subsection{Knowledge distillation}
Knowledge distillation (KD) methods aim to train a student model supervised by a teacher model \cite{hinton2015distilling}. This technique has been widely used for model compression \cite{gou2021knowledge}. The existing  methods can be divided into three categories: response-based methods \cite{zhou2021rethinking,hinton2015distilling}, feature-based \cite{jin2019knowledge,heo2019comprehensive},  and relation-based \cite{tian2019contrastive,tung2019similarity}. 

The response-based methods \cite{zhou2021rethinking,hinton2015distilling} usually utilize the output of the last layer to supervise the training of the student network. The idea of response-based methods is simple and straight-forward. However, they rely on the output of the last layer which cannot supervise the intermediate layers. 

The feature-based methods \cite{jin2019knowledge,heo2019comprehensive} aim to supervise the representation learning of student model by using the intermediate representation as supervision. This idea is first introduced by Fitnets \cite{romero2014fitnets}. Recently, many improvements \cite{chen2021cross,jin2019knowledge} are proposed to import many constraints to reduce the gap between student and teacher. Recently, \cite{yang2021sat} learns auxiliary alignments between 2D object representations and the corresponding objects in 3D scenes to handle visual grounding problem. However, how to choose the favor layers and how to effectively match the feature representation between teacher and student requires further investigation \cite{romero2014fitnets,gou2021knowledge}. 

Different from response-based and feature-based methods, which use the outputs of specific layers in the student model to supervise the student model, the relation-based methods \cite{tian2019contrastive,tung2019similarity} utilize the relationships between different data samples or different layers. A typical example is FPS (flow of solution process) \cite{yim2017gift}, which explores the relationships between different feature maps by calculating their inner products. Lee et al., \cite{lee2018self} use singular value decomposition (SVD) to extract key information in the feature maps to calculate the correlations of feature maps. Peng et al., \cite{peng2019correlation} propose  correlation congruence to utilize both the instance-level information and correlations between instances. Therefore, the student network can learn the instance correlations by using the correlation congruence. However, how to build the relation information from feature maps and data samples is an open question.

Although knowledge distillation has been extensively studied, the main focus was on single tasks and it has not been investigated in the 3D point cloud field, where the knowledge type and distillation method is under-investigated. In this study, we propose a knowledge distillation method to pre-train the 3D backbone and use the pre-trained backbone to solve multiple 3D downstream tasks.

\section{Method}
\begin{figure*}[t!]
    \centering
    \includegraphics[width=0.92\linewidth]{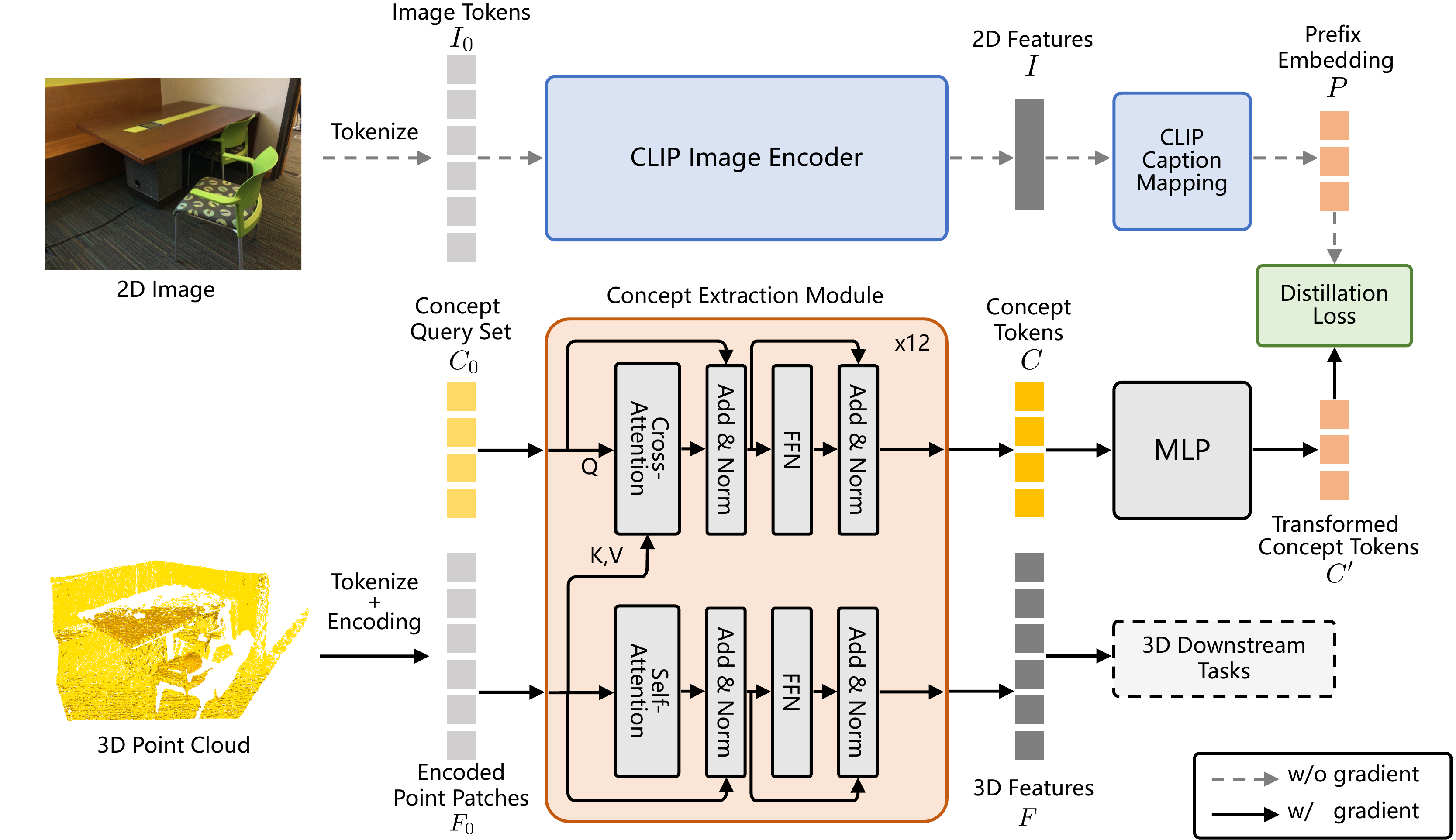}
    \caption{The architecture  of our proposed method for pre-training point cloud backbones through knowledge distillation. There are two main branches for 2D and 3D domains,  respectively. All the parameters from the 2D domain are fixed. 
    A standard transformer structure is implemented as the point cloud backbone while the CLIP image encoder serves as the backbone for 2D images. Concept tokens $C$ are extracted layer by layer from the 3D point cloud backbones through cross attention with a concept query set. By aligning the transformed concept tokens $C'$ with the semantic prefix embedding $P$, knowledge is transferred from the 2D domain to the 3D domain.}
    \label{fig:architecture}
\end{figure*}
We aim to build a knowledge distillation pipeline for 3D point cloud pre-training that can learn from the rich information in the large-scale pre-training model on the image domain. The overall architecture of our model is illustrated in Figure~\ref{fig:architecture}. There are mainly two branches for the point cloud domain and the image domain, respectively. On the point cloud branch, concept information is extracted from a point cloud via the concept extraction module. On the image branch, we adopt a captioning model to generate the semantic prefix embeddings. The two branches are compared by a distillation loss, thus transferring domain knowledge from the image encoder to the point cloud encoder.

In Section \ref{sec:sub1}, we introduce the point cloud encoder and the transformer backbone we adopt. In Section \ref{sec:sub2}, we present the concept extraction network for the point cloud domain. We further explain the extraction of semantic information from the 2D domain in Section \ref{sec:sub3}. Finally, we describe how to keep the alignment between the concept and semantic features in Section \ref{sec:sub4}.

\subsection{Point Cloud Backbone}
\label{sec:sub1}
Following the state-of-the-art self-supervised learning framework PointMAE~\cite{pang2022masked}, the farthest point sampling (FPS) algorithm and k-nearest neighborhood (kNN) algorithm are applied to divide the input point cloud into patches. 
Given an input point cloud with $n$ points $X \in \mathbb{R}^{n*3}$, we first sample the $g$ center points using the FPS algorithm. The $k$ nearest neighbours for each center point are then generated by the kNN algorithm. These local patches are normalized with respect to their center points. This can be formulated as
\begin{align}
    M &= \mathrm{FPS}(X),~(M \in \mathbb{R}^{g * 3}), \\
    T &= \{T_i\}_{i=1}^g = \mathrm{kNN}(X, M),~(T_i \in \mathbb{R}^{k * 3}),
\end{align}
where $M$ represents the set of center points and $\{T_i\}_{i=1}^g$ represents the point cloud patches.

For the point patches, we leverage the PointNet~\cite{qi2017pointnet} to embed them into tokens. This can be formulated as \begin{align}
    F_0 = \mathrm{PointNet}(T),~(F_0 \in \mathbb{R}^{g * d}),
\end{align}
where $F_0$ refers to the embedded tokens and $d$ is the dimension of the embedded tokens. For the positional encoding part, an MLP is developed to encode the position of central points into the embedded dimension $d$.

The point cloud backbone we use here consists of standard transformer blocks. Specifically, we adopt $12$ Transformer Encoder blocks in our experiment. Positional encoding is added to the output layer of each transformer block to add positional information. It can be formulated as
\begin{align}
    F = \mathrm{Transformer(F_0)}, (F \in \mathbb{R}^{g * d}),
\end{align}
where F represents the encoded features. The encoded features can then be applied to downstream tasks, such as classification, detection and segmentation.

\subsection{3D Concept Extraction Module}
\label{sec:sub2}

For the point cloud domain, our goal is to extract a certain type of feature that can be more abstract than the visual features to align with the semantic model. Inspired by a recent work on concept tokenization~\cite{yang2022visual}, we propose to facilitate the alignment by extracting concept tokens from the visual features of point clouds with a concept query set. The concept query set is a collection of learnable concept prototypes. It can be interpreted as the template of common global features for the point clouds. With these learnable concept prototypes, point clouds are encoded into tokens of the concept domain through a cross-attention mechanism. By contrasting concept features with the semantic features of 2D images, we manage to achieve the implicit alignment of features from the point cloud encoder and the image encoder. In other words, the concept domain serves as an intermediate modality between the semantic features of 2D images and the visual features of the 3D point clouds, thus eliminating the domain gap.

The detailed implementation of the Concept Extraction Module is described as follows. We denote the concept query set as $C_0 = \{c_1, c_2, \dots c_{k_{con}}\} \in \mathbb{R}^{k_{con} * d_{con}}$, where $k_{con}$ is the size of concept query set and $d_{con}$ is the dimension for the query tokens. We set $d_{con}$ to be the same dimension as $d$. Based on the concept query set, we design a cross-attention mechanism to extract the concept information from the point cloud. 

The concept extraction module connecting the point cloud domain and concept domain is composed of $N = 12$ cross-attention blocks. For each cross-attention block with $(Q, K, V)$ as input, $Q$ represents the tokens of the concept branch, while $K$ and $V$ are generated from the feature tokens of the point cloud branch.
This can be formulated as \begin{align}
    C_{out}, F_{out} = \mathrm{CrossAttentionModule}(C_{in}, F_{in}),
\end{align}
where CrossAttentionModule consists of a SelfAttention block and a CrossAttention block. $C_{out}, F_{out}$ can be given by \begin{align}
    C_{out} &= \mathrm{CrossAttention}(C_{in}, F_{in}, F_{in}) \\
    F_{out} &= \mathrm{SelfAttention}(F_{in}, F_{in}, F_{in}).
\end{align}
Each cross-attention block or self-attention block is followed by a feed-forward network and a layer normalization. Note that the self attention module here is part of the point cloud backbone we have already explained in Section \ref{sec:sub1}. Thus, we can think of the point cloud backbone as part of the concept extraction module, formulated by\begin{align}
    C, F = \mathrm{ConceptExtractionModule}(C_0, F_0).
\end{align} The concepts are extracted from every layer of the point cloud feature tokens.

\subsection{2D Semantic Module}
\label{sec:sub3}

For the 2D image domain, we aim to generate high-quality supervision for the point cloud domain through the well-pre-trained 2D models. As we have previously explained, a large gap exists between visual features from 2D and 3D domain, so we focus on the semantic features that can be transferable across domains. Due to the success of the hyper-scale pre-trained model CLIP~\cite{radford2021learning} which learns visual concepts from language supervision, it is feasible that we obtain high-quality semantic features directly from 2D images. We take advantage of a recent captioning model ClipCaption~\cite{mokady2021clipcap} to generate the semantic features. 

Given an input 2D image $x \in \mathbb{R}^{d_{\mathrm{img}}}$, we first encode it into the embedded space with a CLIP visual encoder. \begin{align}
    I = \mathrm{CLIP}(x),~(I \in \mathbb{R}^{d_{\mathrm{clip}}}),
\end{align}
where $I$ is the encoded visual features and $d_{\mathrm{clip}}$ is the encoded dimension of the CLIP model. 
The caption model then take the encoded feature $I$ as the input to generate semantic prefix embedding P, \begin{align}
    P = \{p_1, \dots p_l\} = \mathrm{ClipCaption}(I),~(p_i \in \mathbb{R}^{d_{\mathrm{cap}}}).
\end{align} 
The captioning model is well-trained so that the language model can take these prefix embeddings $P = \{p_1 \dots p_l\}$ as input for captioning. That is, the parameters of the pre-trained captioning model $\theta$ should satisfy \begin{align}
    \mathop{\arg\max}_{\theta}\log \mathbb{P}_\theta(c_j\mid p_1,\dots p_l,c_1,\dots,c_{j-1}),
\end{align}
where $c_i$ represents a caption embedding.
Thus, such prefix embedding contains the semantic information we typically need for our goal.

\subsection{Teacher-Student Framework}
\label{sec:sub4}
With all these components explained, we now introduce the teacher-student framework. In our teacher-student paradigm, the backbone of the image domain serves as the teacher model and the backbone of the point cloud domain serves as the student model. For the paired data of a point cloud and a 2D image $(X, x)$, we extract the concept tokens $C \in \mathbb{R}^{k_{con}*d_{con}}$ for $X$ and the semantic prefix tokens $P \in \mathbb{R}^{k_{cap}*d_{cap}}$ for $x$. With a two-layer MLP, $C$ is transformed into $C' \in {k_{cap}*d_{cap}}$ so that it corresponds to the semantic prefix tokens in a shape. A simple yet efficient MSE loss is further implemented to align the transformed concept tokens $C'$ and semantic prefix tokens $P$, which can be formulated as\begin{align}
    \mathcal{L} = \mathrm{MSE} (C', P).
\end{align}

The parameters of the 2D domain are fixed. With back propagation of the gradient, the 3D student model thus gains knowledge from the 2D domain through this framework.
\section{Experiments}

In this section, we describe the experiments that validate our method. We pre-train our model on ScanNetv2, a real-world dataset for indoor scenes. Our model is evaluated on various downstream tasks and compared with the state-of-the-art point cloud pre-training methods. We also conduct ablation studies for every key component of our framework, and demonstrate why we do not adopt reconstruction-based methods.

\subsection{pre-training Setup}
\subsubsection{Dataset Setups}
Our knowledge distillation pre-training is based on the real-world dataset ScanNetV2~\cite{dai2017scannet}, which is an RGB-D video dataset containing 2.5 million views in more than 1500 scans. With a sample frame rate of 25, we gather the paired RGB image data and 3D point cloud data from the scanned RGB-D videos, thus obtaining 100k high-quality data. The data are partitioned into the training set and testing set following the ScanNetV2 Benchmark.
We sample 4096 points from each input point cloud and divide them into 64 point patches by FPS and kNN algorithms we introduced in Section~\ref{sec:sub1}. 

\subsubsection{Pre-training Settings}
We describe the pre-training setting of our backbone.
For the point cloud backbone, we adopt the standard transformer encoder with 12 transformer blocks. The hidden dimension of the transformer blocks is 384, and the number of heads is set to 6 for each block. For the 2D image branch, we leverage the ClipCaption~\cite{mokady2021clipcap} model.

The base learning rate for pre-training is 0.001. We adopt the AdamW optimizer and cosine weight decay scheduler with a weight decay rate of 0.05 for training. 

\subsection{Evaluation on Downstream Tasks}
In this section, we evaluate our point cloud backbones pre-trained by our proposed knowledge distillation methods on various downstream point cloud understanding tasks, including object detection, object segmentation, object classification and part segmentation. Real-world datasets are used in object detection and object segmentation. Synthetic datasets are used in object classification and part segmentation. 

\subsubsection{Object Detection on ScanNetV2}
\label{subsubsec:detection}
Detection on a scene-level downstream task is challenging for 3D models. We evaluate our pre-trained model with the downstream task 3D object detection on ScanNetV2~\cite{dai2017scannet}. ScanNetV2 is a richly annotated real-world dataset built by reconstruction of indoor scenes. There are 1205 training scenes and 312 testing scenes, with instance-level object bounding box labels provided for 18 categories.

We adopt the 3DETR~\cite{misra2021end} model for the downstream task of object detection. 3DETR is an end-to-end transformer model for 3D object detection. It combines the standard transformer structure with non-parametric queries and Fourier positional embeddings to detect the bounding boxes from 3D point clouds. We follow the settings of 3DETR for point sampling and data augmentation. The backbone structure for 3DETR corresponds to the pre-trained model, which contains 12-layer transformer blocks with the transformer dimension of 384.

\begin{table}[t]
  \centering
  \caption{\textbf{Object Detection on ScanNetV2.} We compare the proposed knowledge distillation method with various unsupervised learning methods of VoteNet and 3DETR backbones. Our model show notable performance gains compared with the baseline self-supervised learning methods.} 
  \begin{tabular}{c|cc}
  \shline
    Method & \multicolumn{1}{l}{mAP25} & \multicolumn{1}{l}{mAP50} \bigstrut[b]\\
    \hline
    VoteNet~\cite{qi2019deep} & 58.6 & 33.5 \\
    STRL~\cite{huang2021spatio} & 59.5 & 38.4 \\
    Implicit Autoencoder~\cite{yan2022implicit} & 61.5 & 39.8 \\
    RandomRooms~\cite{rao2021randomrooms} & 61.3 & 36.2 \\
    PointContrast~\cite{xie2020pointcontrast} & 59.2 & 38.0 \\
    DepthContrast~\cite{zhang2021self} & 61.3 & - \\
    \hline
    3DETR~\cite{misra2021end} & 61.1 & 40.2 \\
    Point-BERT~\cite{yu2022point} & 61.0 & 38.3 \\
    Point-MAE~\cite{pang2022masked} & 62.1 & 41.2 \\
    \textbf{Distillation} & \textbf{63.4}  & \textbf{42.2} \\
    \shline
    \end{tabular}%

  \label{table:detection}
\end{table}%

We compare our results with PointBERT~\cite{yu2022point} and PointMAE~\cite{liu2022masked} that are also pre-training methods with 3DETR backbone. The results can be found in Table \ref{table:detection}. From the experiment results, we notice that Point-BERT and Point-MAE only gain less than one point in mAP25 compared to the 3DETR model trained from scratch. This demonstrates that the reconstruction-based self-supervised learning methods only gain limited knowledge for difficult downstream tasks such as object detection. In contrast, our proposed knowledge distillation method improves the mAP25 score by 2.4 points, thus validating the effectiveness of knowledge distillation.

\subsubsection{Semantic Segmentation on ScanNetV2}
Semantic segmentation on a scene-level real-world dataset is a rather challenging task for 3D models, especially for those with transformer backbones that focus on the global features through coarse patching and a self-attention mechanism. Therefore, comparing the transformer backbones on segmentation tasks can clearly demonstrate the quality of pre-trained models. The dataset we use here is still ScanNetV2~\cite{dai2017scannet}. The brief description of ScanNetV2 can be found in Section \ref{subsubsec:detection}. We conduct the semantic segmentation experiments based on the rich semantic voxel labeling on ScanNetV2 scenes.

For the segmentation task, semantic masks are generated by downsampling and upsampling. During downsampling, we randomly select 8192 points from each input point cloud and divide them into 256 point patches. The features of these patches are extracted by our 12-layer transformer encoder. In the upsampling process, we follow the segmentation head of DGCNN~\cite{wang2019dynamic} to select three nearest neighbour center points for each point and concatenate their features. Convolution and normalization are then applied to generate features for predicting segmentation labels. 

The semantic segmentation results can be found in Table~\ref{tab:seg}. From the results, we see that replacing the encoding backbone of DGCNN with transformer causes a large drop in mIOU. The experiment results of Point-MAE and our proposed knowledge distillation method demonstrate that pre-training is essential for the transformer-based models in the downstream task of segmentation. Typically, our method outperforms the state-of-the-art baseline method Point-MAE by 1 point in mIOU.

\begin{table}[t]
  \centering
  \caption{\textbf{Semantic Segmentation on ScanNetV2}. We report the class-level mean IOU for all the compared methods. Our method outperforms the state-of-the-art self-supervised pre-training method Point-MAE by 1 point.}   
  \begin{tabular}{c|c}
  \shline
    Methods & mIOU \\
    \hline
    PointNet++~\cite{qi2017pointnet++} & 33.9 \\
    DGCNN~\cite{wang2019dynamic} & 44.6 \\
    Transformer-DGCNN & 37.0 \\
    \hline
    Point-MAE~\cite{pang2022masked} & 45.4 \\
    \textbf{Distillation} & \textbf{46.4} \\
    \shline
    \end{tabular}%
   \label{tab:seg}
\end{table}%

\subsubsection{Object Classification on ModelNet40}
Object classification is a common vision task. We evaluated the performance of our pre-trained model on ModelNet40~\cite{wu20153d}. ModelNet40 is a popular 3D classification dataset, containing 12311 synthetic CAD objects covering 40 categories. We follow the standard experiment setting to split the ModelNet40 dataset into a training set of size 9843 and testing set of size 2468.  For data augmentation, we implement the generally used random scaling and random translation in the training process. We also adopt the standard voting method.

We compare our results with other supervised learning methods such as standard transformer~\cite{yu2022point} and other self-supervised learning methods such as PointMAE~\cite{pang2022masked}. The object classification experiment results are given in Table~\ref{tab:cls}. Although our knowledge distillation model is pre-trained on a real-world dataset and finetuned on a synthetic dataset, it still gains in accuracy despite the domain gap between these two types of datasets. Our method achieves a 93.0\% accuracy, with an improvement of 0.9\% accuracy compared to the basic transformer without pre-training. For fair comparison, we also pre-train the state-of-the-art self-supervised learning method Point-MAE on ScanNetV2 following all the same pre-training settings as our knowledge distillation methods. Experiments show that we achieve 0.5\% higher accuracy than Point-MAE.


\begin{table}[t]
  \centering
  \caption{\textbf{Object Classification on ModelNet40}. The results are compared with supervised learning methods such as Transformer and self-supervised learning methods such as Point-MAE. Our methods show advantage over the baseline methods pre-trained on ScanNetV2.}
  \begin{tabular}{c|cc}
  \shline
    Methods & pre-train Dataset & Accuracy\\
    \hline
    PointNet~\cite{qi2017pointnet} & - & 89.2 \\
    PointNet++~\cite{qi2017pointnet++} & - & 90.7 \\
    PointCNN~\cite{li2018pointcnn} & - & 92.5 \\
    KPConv~\cite{thomas2019kpconv} & - & 92.9 \\
    DGCNN~\cite{wang2019dynamic} & - & 92.9 \\
    RSCNN~\cite{liu2019relation} & - & 92.9 \\
    Transformer~\cite{yu2022point} & - & 92.1 \\
    \hline
    Transformer-OcCo~\cite{yu2022point} & ShapeNet & 93.0 \\
    Point-BERT~\cite{yu2022point} & ShapeNet & 93.2 \\
    Point-MAE~\cite{pang2022masked} & ShapeNet & \textbf{93.8} \\
    MaskPoint~\cite{liu2022masked} & ShapeNet & \textbf{93.8} \\
    \hline
    Point-MAE & ScanNetV2 & 92.5 \\
    MaskPoint & ScanNetV2 & 92.7 \\
    \textbf{Distillation} & ScanNetV2 & \textbf{93.0} \\
    \shline
    \end{tabular}%

  \label{tab:cls}
\end{table}%


\subsubsection{Part Segmentation on ShapeNetPart}
Object part segmentation requires models to indicate fine-grained object parts instead of only object labels. We conduct the part segmentation experiment on ShapeNetPart~\cite{yi2016scalable}, which covers 16,881 models from 16 categories. Following all the previous works~\cite{pang2022masked, yu2022point, wang2021unsupervised, sharma2020self}, 
we sample 2048 points from each object as input and divide them into 128 point patches. We adopt the same segmentation head with Point-MAE~\cite{pang2022masked}, which selects the features from the 4th, 8th and last layers of the transformer block. After concatenating these three layers of features, we generate two global features by taking average pooling and max pooling operations. Following PointNet++~\cite{qi2017pointnet++}, the concatenated features for the 128 point patches are upsampled back to 2048 points to obtain point-wise features. An MLP is implemented to generate the labels for each point from the extracted two global features.

The results of part segmentation can be found in Table~\ref{tab:partseg}. Our knowledge distillation method is compared with supervised learning methods such as Transformer~\cite{yu2022point} and self-supervised learning methods such as Point-MAE~\cite{pang2022masked}. We report the instance level mean IOU for our method and all baseline methods. Our proposed method gains 0.8 points in mIOU compared with a standard Transformer trained from scratch. Our method and the state-of-the-art self-supervised learning method Point-MAE yield comparable results with pre-training on ScanNetV2 using the same settings.

\begin{table}[t]
  \centering
  \caption{\textbf{Part Segmentation on ShapeNetPart.} We compare our method with all the  baseline methods by the instance level mean IOU. Our method shows comparable results with the state-of-the-art methods.}
  \begin{tabular}{c|cc}
  \shline
    Methods & pre-train Dataset & mIOU\\
    \hline
    PointNet~\cite{qi2017pointnet} & - & 83.7 \\
    PointNet++~\cite{qi2017pointnet++} & - & 85.1 \\
    DGCNN~\cite{wang2019dynamic} & - & 85.2 \\
    Transformer~\cite{yu2022point} & - & 85.1 \\
    \hline
    Point-BERT~\cite{yu2022point} & ShapeNet & 85.6 \\
    Point-MAE~\cite{pang2022masked} & ShapeNet & \textbf{86.1} \\
    \hline
    Point-MAE & ScanNetV2 & \textbf{85.9} \\
    \textbf{Distillation} & ScanNetV2 & \textbf{85.9} \\
    \shline
    \end{tabular}%

  \label{tab:partseg}%
\end{table}%

\subsection{Ablation Studies}

\subsubsection{Effectiveness of Key Components}
In this section, we study the effectiveness of the main components of our proposed method.
We first show the ablation study results for our proposed method on the downstream task of classification on ModelNet40.
Two main components for our methods have been evaluated: the semantic captioning module and the concept extraction module. Furthermore, we implement a method of directly aligning the 2D and 3D features by semantic alignment. We also compare our knowledge distillation results with the same backbone trained from scratch.
We do not adopt the voting strategy in ablation studies.

Table \ref{tab:ablation} is a summary of the ablation study of the components of our network. 
The row "\textbf{from scratch}" shows the result of directly training a standard 12-layer transformer without any pre-training. 
The row "\textbf{w/o cross attention}" represents the experiment where we remove the cross attention module for concept extraction. Instead, we connect the MLP layer for concept alignment to the self-attention blocks to align the extracted features $F$ with the semantic prefix $P$. In other words, the distillation loss $\mathcal{L}$ is modified to $\mathrm{MSE}(F', P)$. 
The row "\textbf{w/o caption model}" refers to the pre-training method of aligning concept with the CLIP encoded features instead of the semantic prefix captions. 
The row "\textbf{direct local alignment}" refers to the method of directly aligning the local features from 2D and 3D domains. More specifically, we compare each encoded patch from the 2D domain with the encoded patch from 3D domain of the nearest projection distance. The direct alignment method can be considered as removing both concept extraction and caption modeling from our methods, but provides a more interpretable way of contrasting in feature alignment.
The row "\textbf{distillation}" refers to our proposed method of knowledge distillation through concept alignment.

\begin{table}[t]
  \centering
  \caption{Ablation Studies of the classification downstream task on ModelNet40 without the voting strategy for our proposed knowledge distillation method.} 
  \begin{tabular}{c|c}
  \shline
       Methods  & Accuracy\\
    \hline
    From Scratch &  92.38 \\
    w/o Concept Extraction & 92.42 \\
    w/o Caption Model & 92.18 \\
    Direct Local Alignment & 91.65 \\
    Distillation & 92.71 \\
    \shline
    \end{tabular}%

  \label{tab:ablation}%
\end{table}%

The results show that every component of our proposed method is essential. From the results, we see that removing the captioning model from our model results in a 0.53 loss in accuracy, even worse than training from scratch. This supports our initial claim that domain gap is a serious problem in knowledge distillation between 2D and 3D domains. Removing concept extraction from our model still results in a 0.29 loss in accuracy. This shows that it is difficult for the 3D point cloud backbones to directly learn from contrasting with semantic prefixes. Our proposed concept alignment mechanism plays an essential role in bridging the domains between 2D semantics and 3D features. It is interesting to find out that the direct alignment of local patches does not typically work for the object classification task. This strengthens our argument that concept alignment mechanism is necessary in the knowledge distillation across 2D and 3D domains.

\subsubsection{Do We Need Reconstruction?}
Currently, most of the state-of-the-art 3D self-supervised learning methods for transformer backbones rely on the reconstruction mechanism. It is straightforward  to think of adding the reconstruction mechanism to our proposed pipeline. In this subsection, we present the experiment results regarding adding a reconstruction module to our proposed knowledge distillation backbone.

The experiments are conducted on the same downstream task of classification on ModelNet40. We implement a simple reconstruction module for the 3D point cloud encoder to compare the experiment results with regard to reconstruction. 
More specifically, we mask 60\% of the encoded point cloud features, and try to reconstruct them through a 3-layer transformer decoder. We recover the coordinates of all the masked points and compared them with the ground truth points. The reconstruction loss is calculated by the $l_2$ Chamfer Distance. We compare the results of the models trained  with only the  reconstruction loss, trained only with the distillation loss, and trained with both losses.

\begin{table}[t]
  \centering
  \caption{Experiments results regarding the reconstruction for the classification downstream task on ModelNet40.}
  \begin{tabular}{c|c}
  \shline
       Methods  & Accuracy\\
    \hline
    Only Reconstruction & 92.42 \\
    Only Distillation & 92.71 \\
    Reconstruction + Distillation & 92.42 \\
    \shline
    \end{tabular}%
  \label{tab:reconstruction}%
\end{table}%

The results can be found in Table~\ref{tab:reconstruction}. From the comparison between the pretrianed models with the reconstruction loss only and with the distillation loss only, we see that the distillation-based method shows some advantages over the simple reconstruction-based method. However, adding these two losses together will not boost the performance. Instead, the overall performance is constrained by the reconstruction loss.
The reason is that the reconstruction task is easier than the concept alignment task, thus can be learned more easily by encoder backbones.
Combining these two losses can cause the pre-trained model to easily converge to a local optimum due to the reconstruction loss.
Therefore, the distillation loss is a more sophisticated loss compared with the reconstruction loss.

\begin{table}[t]
  \centering
  \caption{The training reconstruction loss after 250 epochs using different training losses.}
  \begin{tabular}{c|c}
  \shline
        Training method & Recon-loss@epoch250 \\
    \hline
        No Training   &  1.10 \\
        Only Reconstruction & 3.95e-4 \\
        Only Distillation & 5.63e-3 \\
    \shline
    \end{tabular}%

  \label{tab:loss}%
\end{table}%

In fact, training a 3D backbone with only the distillation loss can also enhance the model's capability of reconstructing 3D point clouds. In Table~\ref{tab:loss}, we present the average training reconstruction loss for different models after 250 epochs of pre-training on ScanNetV2. The first row "bo training" represents the initial reconstruction loss with randomly initialized parameters. In the experiment of "only reconstruction", both the transformer encoder and the transformer decoder are trained with the reconstruction loss. In the experiment of "distillation", only the transformer encoder is trained using the distillation loss. The experiment results show that by training only with the distillation loss, the reconstruction loss can decrease to a certain extent even with a random initialized decoder. This demonstrates that the distillation method can indeed train a high-quality transformer encoder without reconstruction techniques.


\section{Conclusion}
We present a knowledge distillation framework for 3D point cloud backbones to directly learn from well-trained 2D pre-trained model CLIP. We address the problem of domain gap by a concept alignment mechanism. Our pre-trained model achieves significant improvements for the transformer backbones on a variety of downstream tasks, such as object detection, semantic segmentation, object classification and part segmentation. Extensive experiments on both synthetic datasets and real-world datasets demonstrate the effectiveness of our knowledge distillation method. Knowledge distillation across 2D and 3D domains is a subject worth further investigation. We hope our work in this field can spawn the research of knowledge distillation across 2D and 3D domains.

{\small
\bibliographystyle{ieee_fullname}
\bibliography{main}
}

\appendix
\section{Implementation Details}
In this section, we give a more detailed description for all the implementations of our experiments.

\subsection{The Pre-training Experiment}
The knowledge distillation framework basically contains a point cloud branch and a 2D image branch. The detailed information our framework is as follows. For the point cloud backbone, our model basically contains a point cloud tokenizer, a set of self-attention blocks and a set of cross-attention blocks. For the 2D image backbone, we follow the structure of Clip Caption model. We conduct all the pre-training experiments on the ScanNetV2 dataset~\cite{dai2017scannet} with a video sample frame rate of 25.

\textbf{Point Cloud Tokenizer:} Each input point cloud is first downsampled to 4096 points. We then divide them into 64 point patches, each containing 32 points. 
Here, we do not cover all the downsampled points for two reasons. First, there are redundant points in a real-world dataset. Second, we want our model to be capable of extracting the concept information from partial input point clouds.
After grouping into point patches, we implement a point patch encoder with 1-D convolution layers. The output dimension of these 1-D convolution layers are 128, 256, 512, 384. We adopted the batch normalization and ReLu activation layers between these convolutions.

\textbf{Self-attention Blocks:} The self-attention blocks in our backbone follow the standard Transformer architecture for fair comparisons\cite{yu2022point}. Each self-attention block is composed of a multi-head self-attention layer and a FeedForward Network. The multi-head attention can be formulated as\begin{align}
    \mathrm{MultiHead}(Q,K,V) = [\mathrm{head}_1, \dots, \mathrm{head}_{12}] W^O,
\end{align}
where $\mathrm{head}_i = \mathrm{Attention}(QW_i^Q, KW_i^K, VW_i^V)$.
We apply the layer normalization for these two layers. Our transformer encoder consists of 12 self-attention blocks. The hidden dimension of the transformer blocks is 384, and the number of heads is set to 6 for each block.

\textbf{Cross-attention Blocks:} The cross-attention blocks in our backbone share the same parameters so that we can calculate the cross-attention for the features extracted from the self-attention module layer by layer.  dimension 384 and depth 12. The number of concept tokens we choose here is 32, which means we have 32 learnable concept query tokens in our model. After the cross-attention blocks, the concept tokens are extracted. They are then sent into an MLP to align with the semantic prefix embeddings.

\begin{table}[t]
  \centering
  \caption{Experiment settings for our knowledge-distillation framework pre-training.}
  \begin{tabular}{c|c}
    \shline
       \textbf{Config}  & \textbf{Value} \\
    \hline
    Optimizer & AdamW \\
    Learning Rate & 1e-3 \\
    Weight Decay & 5e-2 \\
    Learning Rate Scheduler & cosine \\
    Epoch & 250 \\
    Warmup epoch & 10 \\
    Augmentation & scale and translate \\
    Number of Points & 4096 \\
    Patch Size & 32 \\
    Number of Patches & 64 \\
    Number of Concept Tokens & 32 \\
    Encoder Dim & 384 \\
    Transformer Dim & 384 \\
    Concept Dim & 384 \\
    Encoder Depth & 12 \\
    Number of Heads & 6 \\
    2D Caption Model & Conceptual Captions \\
    Semantic Prefix Length & 10 \\
    Dataset & ScanNetV2 \\
    Sample Frame Rate & 25 \\
    \shline
    \end{tabular}%
  \label{tab:parameters}%
\end{table}%

\textbf{2D Backbone:} For the 2D image branch, we leverage the ClipCaption model pre-trained on Conceptual Captions dataset. The Clip model for the 2D backbone is "ViT-B/32". The semantic prefix length generated by the ClipCaption model is 10.

\textbf{Optimization Setting:} The base learning rate for pre-training is 0.001. We adopt the AdamW optimizer and cosine weight decay scheduler with a weight decay rate of 0.05 for training. The learning rate schedular is a simple Consine scheduler with 10 epochs of warm-up period. The data augmentation we use here is simply the scale and translate.

All the settings of our pre-training experiment have been concluded in Table~\ref{tab:parameters}. We reclaim here that, the pre-train of baseline method Point-MAE~\cite{pang2022masked} follows exactly the same setting as our knowledge distillation model for fair comparison.

\subsection{Settings of Downstream tasks}
We now explain the settings of downstream tasks. In the downstream tasks, we make use of the self-attention backbone of our knowledge distillation framework as the transformer backbone, thus being a 12-layer standard transformer. The cross-attention part and the 2D image part, which are designed for knowledge distillation, are discarded in the downstream tasks. The other details of downstream tasks can be found as follows.

\textbf{Object Detection:} In the object detection downstream task, we adopt the 3DETR~\cite{misra2021end} model. The basic idea of 3DETR is to combine the standard transformer structure with non-parametric queries and Fourier positional embeddings to detect the bounding boxes from 3D point clouds. We follow the settings of 3DETR that pick 2048 points for each point cloud and generate 256 queries. The encoder structure corresponds to our knowledge distillation pre-trained model, which contains a 12-layer transformer blocks with the transformer dimension of 384. The decoder is a 8-layer transformer decoder with dimension 256. We finetune the model with base learning rate 5e-4 for 720 epochs.

\textbf{Semantic Segmentation:} In the semantic segmentation downstream task, semantic masks are generated by downsampling (or patching) and upsampling. The number of points in input point clouds in 8192. We divide them into 256 point patches, each with the group size of 32. In the upsampling process, we follow the segmentation head of DGCNN~\cite{wang2019dynamic} to select three nearest neighbour center points for each point and concatenate their features. Two 1D convolution blocks are applied to generate the point-wise predictions for the 20 prediction classes. We adopt the Adam Optimizer, with base learning rate of 0.001 and weight decay rate of 0.01. The model is finetuned for 200 epochs.

\textbf{Object Classification:} In the object classification downstream task, we follow the standard experiment setting to split the ModelNet40 dataset into a training set of size 9843 and testing set of size 2468. The number of points in the input point clouds is 1024. They are patched into 64 groups with the group size of 32 during encoding. For data augmentation, we implement the generally used random scaling and random translation in the training process. We also adopt the standard voting method. The base learning rate is set to 0.0005 with the weight decay rate of 0.05, using AdamW optimizer. The model is finetuned for 300 epochs.

\textbf{Part Segmentation:} In the part segmentation downstream task, we follow the settings of previous works~\cite{pang2022masked, yu2022point, wang2021unsupervised, sharma2020self}. We sample 2048 points from each input, dividing them into 128 point patches with size 32. Following the segmentation head of Point-MAE~\cite{pang2022masked}, we select the features from the 4th, 8th and last layers of the self-attention block and concatenate them. Two global features are then generated by average pooling and max pooling operations. The concatenated features for the 128 point patches are upsampled back to 2048 points with the PointNet++~\cite{qi2017pointnet++} decoder to obtain point-wise features. We finentune the model for 300 epochs, with base learning rate of 0.0002.

\section{More Experimental Results}
In this section, we give more detailed experimental results with visualizations for one of our downstream tasks: object detection on ScanNetV2.

\begin{figure*}[]
    \centering
    \includegraphics[width=0.7\linewidth]{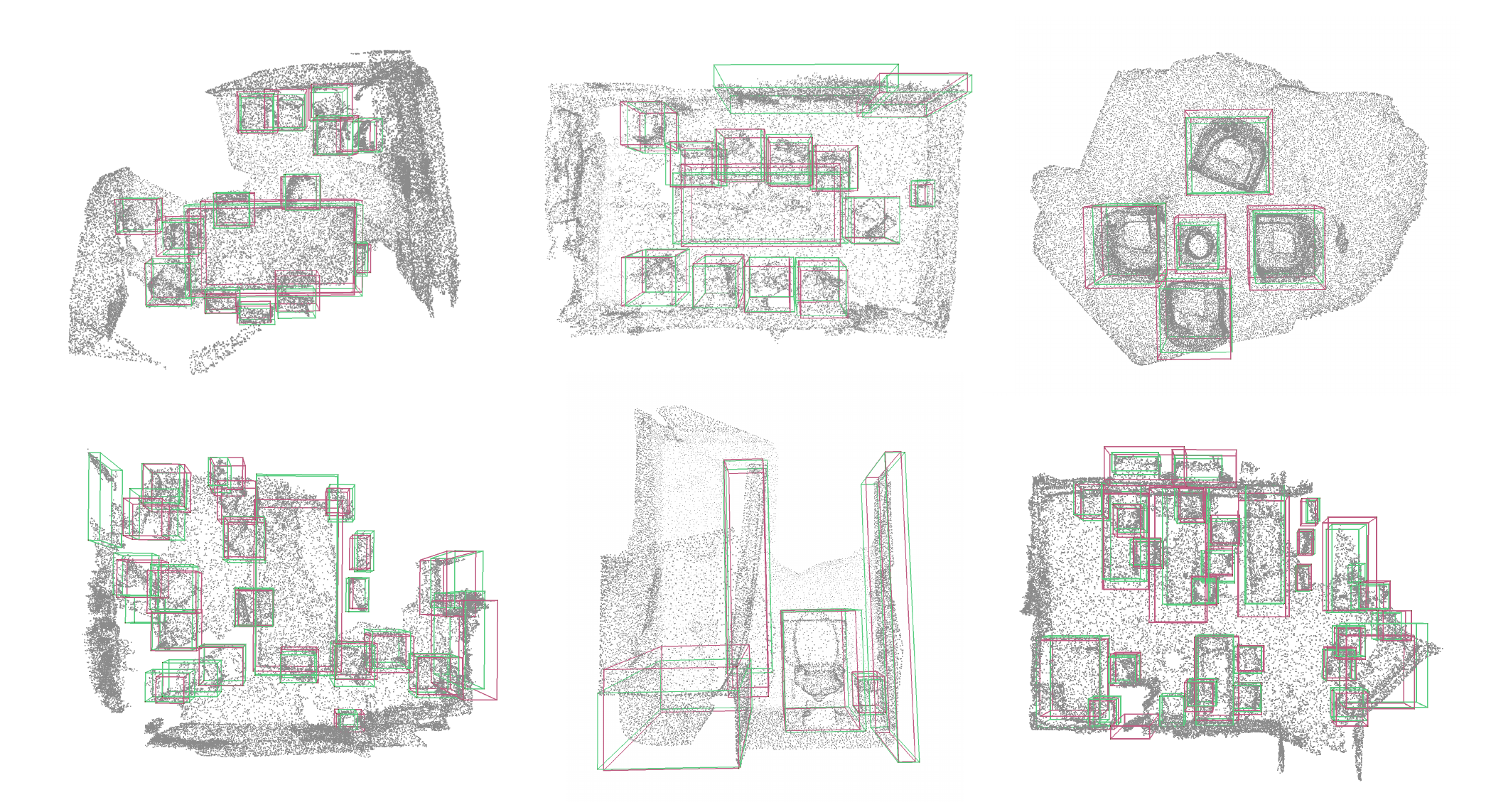}
    \caption{The detection results on ScanNetV2. The green bounding boxes are the ground truth boxes, and the red bounding boxes are the predicted boxes of our proposed knowledge distillation method. The visualizations illustrate that our distilled model can generate high-quality detection results.}
    \label{fig:det-visualize}
\end{figure*}

Table~\ref{tab:per-class-det} presents a result of the per-class $\mathrm{AP}_{25}$ for the detection task on ScanNetV2. We present the comparison between our proposed knowledge distillation method and the state-of-the-art self-supervised learning method Point-MAE. From the results, we see that our proposed distillation method shows comparable results with Point-MAE in most classes. However, for some particular classes such as bookshelf, curtain and shower curtain, the proposed distillation method outperforms the baseline method by more than 6 points in $\mathrm{AP}_{25}$. This can be attributed to the effects of knowledge distillation, \ie the features of these classes have been learned from multiple domains (2D + point cloud), thus boosting the performance.

We also provide visualizations for the object detection task on ScanNetV2, which can be found in Figure~\ref{fig:det-visualize}. From the visualizations, we clearly see that our proposed model can generate high-quality detection results.

\begin{table*}[t]
\small
\setlength\tabcolsep{2.5px}
  \centering
  \caption{Per-class $\mathrm{AP}_{25}$ for ScanNetV2.} 
  \begin{tabular}{c|cccccccccccccccccc}
  \shline
    Model & cabin & bed & chair & sofa & table & door & windo & b-shelf & pic & counter & desk & curtain & refrig & sh-curt & toilet & sink & b-tub & g-bin\\
    \hline
    Point-MAE & 48.6 & 81.2 & 88.0 & 90.6 & 66.9 & 50.9 & 40.0 & 43.8 & 12.7 & 55.1 & 70.7 & 53.8 & 55.4 & 59.7 & 99.6 & 68.3 & 84.1 & 47.2 \\
    Distillation & 45.4 & 80.0 & 87.6 & 87.8 & 63.8 & 49.9 & 40.0 & 50.6 & 12.4 & 57.6 & 72.8 & 60.4 & 61.0 & 67.2 & 99.2 & 74.8 & 84.2 & 46.6 \\
    \shline
  \end{tabular}%

  \label{tab:per-class-det}%
\end{table*}%

\section{More Ablation Studies}
In this part, we give another ablation study about the way of sampling. For our proposed architecture, the running time determines largely on the number of points in a downsampled point cloud, the number of point patches and the number of points encoded in each patch. Therefore, it is not feasible that we choose all these three parameters with large number. As is discussed in our pre-training experiment settings, we choose 4096 points for our input point clouds, and the grouping parameters of 64 patches with patch size 32. Not all the downsampled points have been covered in this setting. In this part, we study the effects of the settings of downsampling and patching.

\begin{table}[t]
  \centering
  \caption{Ablation Studies of the classification downstream task on ModelNet40 regarding downsampling and patching.} 
  \begin{tabular}{ccc|c}
  \shline
    Num Points & Num Patches & Patch Size  & Accuracy\\
    \hline
    1024 & 64 & 32 & 92.02 \\
    4096 & 64 & 32 & 92.71 \\
    \shline
  \end{tabular}%

  \label{tab:sample}%
\end{table}%

We experimented on the way of downsampling and patching in the pre-training process for our knowledge distillation framework, and compared the results on the downstream task, \ie object classification on ModelNet40. No voting techniques are used in the experiment. The results for this ablation study can be found in Table~\ref{tab:sample}. From the table, we see the three settings of downsampling and patching. The setting on the first row is the setting which the baseline method PointMAE adopts when pre-training on the synthetic dataset ShapeNet. The setting on the second row is what our pre-trained knowledge distillation model uses. 

\begin{figure}
    \centering
    \includegraphics[width=\linewidth]{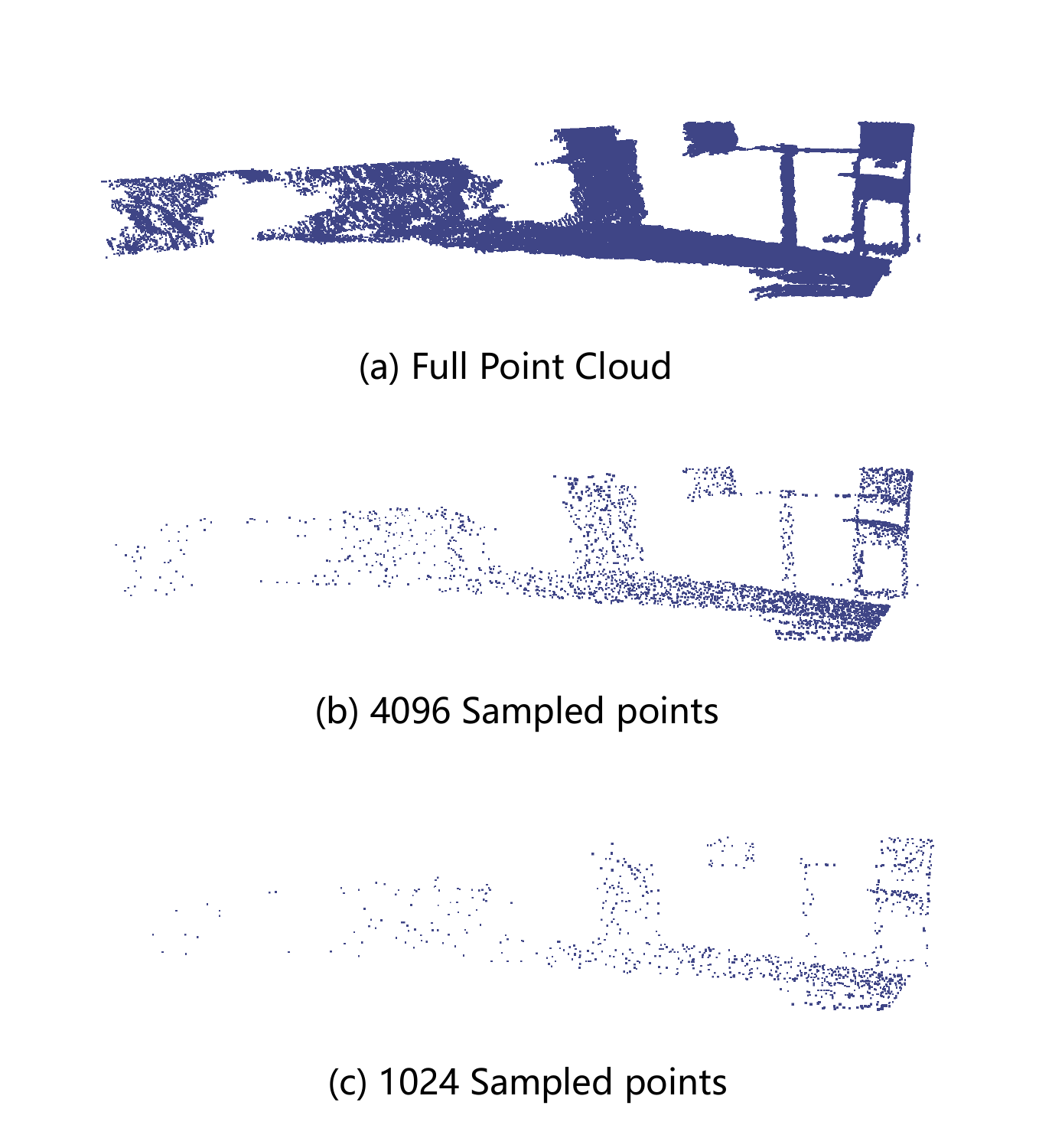}
    \caption{We compare the quality of point clouds under different number of downsampled points. With number of points set to 4096, the original scene is still clear. When the number of points decreases to 1024, the point cloud evidently loses information.}
    \label{fig:sample}
\end{figure}

Here we analyze the results. From the comparisons between the two settings, we see that the common way of sampling and patching on synthetic datasets does not work well on real-world dataset ScanNetV2. We further conclude that the main difference for these two settings is that the first setting tries to sample less points to ensure a better coverage of points by the patches. The better performance of the second setting demonstrates that the quality (or number of points) of the sampled point clouds is more important than the complete coverage in point cloud downsampling and patching.

Why is it essential to keep a large number of sampled points in real-world datasets? The reason is that real-world dataset needs much more points to depict the whole scene. Figure~\ref{fig:sample} illustrates the comparison of the quality of point clouds under different number of downsampled points. We see that with the number of points set to 4096, the point cloud still clearly depict the original scene of full point cloud. However, when the number of points decrease to 1024, even humans cannot well understand the scene. Therefore, choosing the number of input points to be 1024 will cause the model to overfit the limited number of points, leading to low-quality pre-training.

\end{document}